\documentclass[compsoc]{IEEEtran}

\usepackage[english]{babel}
\usepackage[utf8]{inputenc}
\usepackage[T1]{fontenc}
\usepackage[citecolor=black,
            colorlinks=true,
            breaklinks]{hyperref}
\hypersetup{%
  pdfinfo={%
    Title={Determining Song Similarity via Machine Learning Techniques and Tagging Information},
    Keywords={Machine Learning, Collaborative filtering, Music, Similarity}%
  }
}
\usepackage{fullpage}
\usepackage[square]{natbib}
\usepackage{amsmath,amssymb,amsfonts,textcomp}
\usepackage{microtype}
\usepackage{booktabs}
\usepackage{balance}
\usepackage{graphicx}

\input{glyphtounicode}

\usepackage{multicol}

\title{Determining Song Similarity via Machine Learning Techniques and Tagging Information}
\author{%
\IEEEauthorblockN{%
Renato L. F. Cunha\IEEEauthorrefmark{1}\IEEEauthorrefmark{2},
Evandro Caldeira\IEEEauthorrefmark{1}}
Luciana Fujii\IEEEauthorrefmark{1},
\begin{multicols}{2}
\IEEEauthorblockA{\IEEEauthorrefmark{1}Universidade Federal de Minas Gerais\\
PPGCC\\
\columnbreak%
\IEEEauthorrefmark{2}IBM Research\\
Brazil Research Lab}
\end{multicols}\vspace*{-1cm}
}

\begin{document}

\maketitle

\begin{abstract}
    The task of determining item similarity is a crucial one in a recommender
    system. This constitutes the base upon which the recommender system will
    work to determine which items are more likely to be enjoyed by a user,
    resulting in more user engagement. In this paper we tackle the problem of
    determining song similarity based solely on song metadata (such as the
    performer, and song title) and on tags contributed by users. We evaluate our
    approach under a series of different machine learning algorithms. We
    conclude that tf-idf achieves better results than Word2Vec to model the
    dataset to feature vectors. We also conclude that \emph{k}-NN models have
    better performance than SVMs and Linear Regression for this problem.
\end{abstract}

\section{Problem description}

Various recommender systems use a metric known as song similarity to predict
candidate songs users would be interested in listening to. Defining such
a metric is somewhat subjective, though, and researchers use two different
approaches for this:
\begin{itemize}
    \item The objective approach, in which similarity is based on
        content information, such as spectral or rhythmic analysis of songs, and the
    \item subjective approach, in which user-generated data, such as tags---also
        known as collaborative filtering---is used.
\end{itemize}

In this project we intend to use the subjective approach to define song
similarity. In particular, we will define the similarity level between two
songs ranging from zero (completely dissimilar) to one (identical) and will
compute it using the co-occurrences of pairs of items in users' histories using
the cosine metric. This metric will also be our model of reality and,
therefore, our ground truth. Such definition is plausible, since researchers
of the field have used it with success~\citep{linden2003amazon}.


\section{Data}

The dataset used in this project was generated by calling
Last.fm's™\footnote{\url{http://last.fm}--Last.fm is a trademark by Audioscrobbler Limited.} API and persisting the results. It
contains more than 5M songs with all associated metadata (tags, artist, album,
play count, number of listeners, duration, mbid\footnote{MusicBrainz ID---a
reliable and unambiguous identifier in the MusicBrainz database
(\url{musicbrainz.org}).}), the listening history of 380K users, and similarity
metrics for 138M pair of songs in our dataset.

A lot of Last.fm™'s data is uploaded by users, for instance, users define tags
for a song. The dataset contains tags that are written in different forms such
as causing inconsistencies and different hyphenation or symbols (e.g.\ Guns \&
Roses \emph{versus} Guns N' Roses) duplicated songs and other noise forms that
we will have to pre-process to achieve better results.

During collection, the data was stored in a MongoDB database, where each API
response was stored as a different JSON document in the database.
Figure~\ref{fig:samplejson} shows an example of such a format. Its fields are:

\begin{itemize}
    \item name: The song's name;
    \item tags: An array of pairs consisting of (name, count), where
                 ``name'' is a tag defined by a user and ``count'' represents
                 how many users have applied that tag to that song. Notice that
                 ``count'' is capped to 100.
    \item album\_mbid: The unique MusicBrainz ID assigned to the album that
                 contains this particular song;
    \item artist\_name: The name of the artist that recorded this particular
                 song;
    \item mbid: The song's unique MusicBrainz ID;\
    \item album\_title: The title of the album that contains this song;
    \item artist\_mbid: The artist's MusicBrainz ID.\
\end{itemize}

\begin{figure}
    \scriptsize
    \begin{verbatim}
{
  'name': 'headspin', 'tags': [['idm', 100],
  ['electronic', 54], /* more tags */],
  'album_mbid': 'a960877b-0319-48ce-8658-c17b1e0dab9a',
  'artist_name': 'plaid',
  'mbid': '3e34ad31-8fd2-4c6c-95a7-7c1fe2bb3dbf',
  'album_title': 'not for threes',
  'artist_mbid': '7e54d133-2525-4bc0-ae94-65584145a386'
}
    \end{verbatim}
    \caption{Sample output of the last.fm™ API as stored in our
    database.}\label{fig:samplejson}
\end{figure}

Additionally, the computed co-occurrence model was computed between song pairs
and stored in a comma-separated file in the format
$(\text{song1},\text{song2},\text{similarity})$, which we had to parse to
correctly build the similarity graph.

\subsection{Data transformation}\label{sec:data}

To correctly model the data we needed to process it in different steps: after
data collection, we had to extract data from MongoDB\footnote{Which we learned
not always returns all documents matching a query.}, normalize text and
integrate the similarity calculations into this data. Data normalization
consisted of removing accents and all kinds of special characters from words,
replacing numbers with words and converting Unicode characters to the closest
latin characters that represented them. All strings in the dataset were
normalized; namely: album title, artist name, song name, and tag name. Since MBIDs are
unique, those were converted to integers sequentially in the order they
appeared.

Once all data was converted, we proceeded to create the feature vectors,
which were created with using two different models: Word2Vec and tf-idf,
described in the following. We also evaluate the various algorithms by
artificially filtering songs with too low similarity: we produced two new
datasets, one in which no songs with similarity smaller than $1\%$ are found,
and another in which no songs with similarity smaller than $2.5\%$ are found.

\subsubsection{Word2Vec}

Word2Vec~\cite{mikolov2013efficient} is a group of models for computing
continuous vector representations of words from very large datasets, and is
particularly well suited for Natural Language Processing (NLP) tasks,
particularly because word vectors are positioned in the vector space such that
words that share common contexts in the corpus are located in proximity
to one another in the space. Therefore, we decided it would be appropriate to
use such a model for defining feature vectors.

We used all the text columns to construct our Word2Vec model. We created
vectors of length 100 and inspected the model manually to see if it was
representative of what we expected. Text with multiple words was considered as
one word only, for instance, ``Rolling Stones'' became one word
``rollingstones'' instead of two separate words ``rolling'' and ``stones''.
Some similarity examples are shown in Table~\ref{tab:word2vec-similarity}.

\begin{table}
    \caption{Examples obtained while doing manual inspection of the generated
    Word2Vec model.}\label{tab:word2vec-similarity}
    \begin{center}
    \begin{tabular}{lll}\\
        \toprule
        \textbf{word 1} & \textbf{word 2} & \textbf{similarity}\\
        \cmidrule(r){1-3}
        samba & bossa & 0.66213981365716956\\
        electronic & techno & 0.83948800290761028\\
        \cmidrule(r){1-3}
        Vocabulary size & \multicolumn{2}{c}{1683231} \\
        \bottomrule
    \end{tabular}
    \end{center}
\end{table}

From the Word2Vec model, we created a feature vector for each song using the
weighted average of the tag vectors, where the weight of each tag was its tag
count for that song, plus the artist vector with a weight of 100, which is the
maximum tag count.

\subsubsection{Tf-idf}

We also modeled features using a term frequency–inverse document frequency
(Tf-idf) model: we decided to treat each song's set of tags as a different
document and constructed a bag-of-words model for the whole dataset, in which
each song was a different document. So the feature set now would be the term
frequencies of each word.

Since we had many tags, these features had to be, initially, represented as a
sparse matrix. We exploited the tag frequency information provided by the
last.fm™ API to build a more correct model: each tag was repeated $n$ times,
where $n$ is the tag count obtained by the last\@.fm API\@.

Also, we tried to increase the weights of less frequent tags by also using the
inverse document-frequency weighting technique. Once the set of features was
determined, we applied Single Value Decomposition (SVD) for feature
decomposition and dimensionality reduction to go from 5000 features (from the
tf-idf model) to 100 (the number of components specified in the SVD).

\subsection{Feature matrix construction}

Since we had too many songs to fit in a reasonably-sized computer's RAM, we
were forced to work on a subset of all songs. We also had to make sure that the
resulting matrix made sense. Therefore, instead of simply slicing the dataset,
we constructed an adjacency list graph representation of all the songs for
which we had some similarity information. Then, we traversed this graph
extracting features to build said matrices. Hence, in the feature matrix we
had, for each pair of songs (up to a limit), we had from columns $1$ to $m$ the
features from the first song of the pair, and from columns $m+1$ to $n$ we had
the features of the second song of the pair (where $m = n/2$ is the number of
features of a song). In the $y$ vector the corresponding line had the
similarity value of both songs. When fed into the models, the X matrix was
further transformed to have only $m$ columns by subtracting the first $m$
columns by the second $m$ columns.

\section{Methodology: Proposed solution \& Algorithms}

We want to be able to predict the similarity between two songs when we have no
co-occurrence data for them, for instance, for when a new song debuts. We will
split the information we have about songs and their similarities item-to-item
into training and test sets and will try to find a model that can
predict similarity without using user play history. Note that the similarity
metric we have right now was computed using only users' history, from which we
derived the co-occurrences between songs, but no other metadata.

\begin{figure}
    \includegraphics[width=.5\textwidth]{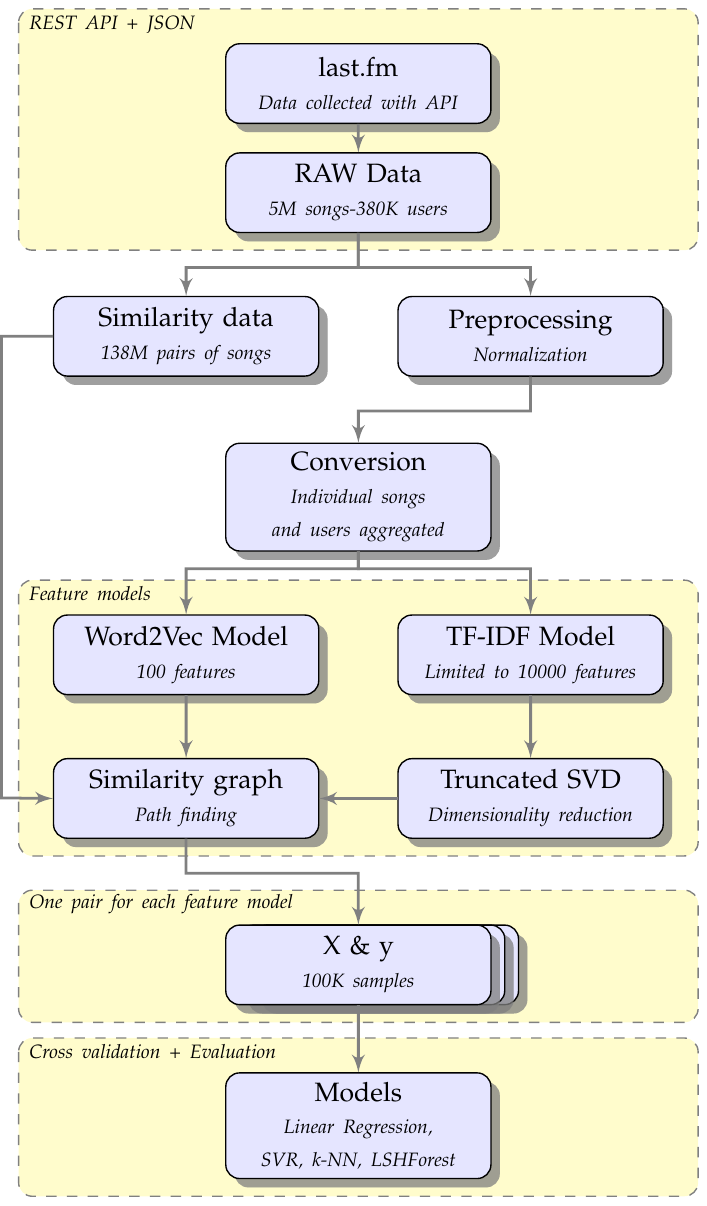}
    \caption{Flow of the data using the methodology described in this
    work.}\label{fig:methodology}
\end{figure}

Figure~\ref{fig:methodology} outlines how the data flows from last.fm™, the
transformations we performed and how the features and labels were obtained from
the data for training the models.

\subsection{Algorithms}

We have evaluated our engineered features with the following models: Linear
Regression (LR), Support Vector Machines Regression (SVR)\footnote{An
adaptation of Support Vector Machines for regression
problems.}~\citep{smola2004tutorial,gunn1998support}, exact and approximate (by
means of locality sensitive hashing~\citep{andoni2006near,bawa2005lsh})
\emph{k}-Nearest Neighbors (\emph{k}-NN) with kernel
regression~\citep{terrell1992variable} to predict new song similarity scores.

%

Since none of these algorithms work directly with the data returned by the
last.fm™ API, we have transformed the data as described in
Section~\ref{sec:data}.

As aforementioned, we have used the cosine metric to measure the similarity
between songs. Therefore, we define the metric here for completeness. Given two
vectors $\vec{x}$ and $\vec{y}$, the similarity between them is defined as the
function
\[
    \mathrm{sim}(\vec{x}, \vec{y}) = \frac{\sum_i x_i y_i}{\sqrt{\sum_i x^2_i}\sqrt{\sum_i y^2_i}}
\]
and, since our feature vectors are composed of non-negative real numbers and
without degenerate cases such as vectors with norm equal to zero, this metric
will only return values between zero and one.

\subsection{Feature Scaling}

Feature scaling is needed when using SVM models, as can be confirmed by
observing Tables~\ref{tab:svm-feature-scaling} and~%
\ref{tab:svm-feature-scaling-tfidf}. In that table we see the $R^2$ score of
how the model performed in the test set for different filtering values in the
features and with raw and scaled ($\mu=0$ and $\sigma^2=1$). Due to that, we
decided to make the whole input have $\mu=0$ and $\sigma^2=1$. The parameters
obtained for such scaling were done only over the training set, since in
practice we will never have access to the whole dataset. Prior to testing the
algorithms, though, we used the same parameters found in the training set to
scale the test set.

\begin{table}
    \caption{Effects of feature scaling on the SVM model when executed with the
    Word2Vec data. Values shown were obtained by running the best
    cross-validated model against the test set
    data.}\label{tab:svm-feature-scaling}
    \begin{center}
    \begin{tabular}{lccc}
        \toprule
        & \multicolumn{3}{c}{$R^2$ score}\\
                         & No filtering & similarity > 0.01 & similarity > 0.025 \\
        \cmidrule(r){2-4}
        Raw              & -27.547      & -0.825            & -0.302 \\
        Scaled           & -24.610      & -0.764            & -0.170 \\
        \bottomrule
    \end{tabular}
    \end{center}
\end{table}

\begin{table}
    \caption{Effects of feature scaling on the SVM model when executed with the
    tf-idf model. Values shown were obtained by running the best
    cross-validated model against the test set
    data.}\label{tab:svm-feature-scaling-tfidf}
    \begin{center}
    \begin{tabular}{lccc}
        \toprule
        & \multicolumn{3}{c}{$R^2$ score}\\
                         & No filtering & similarity > 0.01 & similarity > 0.025 \\
        \cmidrule(r){2-4}
        Raw              & -13.225      & -0.725            & -0.164 \\
        Scaled           & -4.566       & -0.490            & -0.156 \\
        \bottomrule
    \end{tabular}
    \end{center}
\end{table}

\subsection{Support Vector Machine Regression}

We implemented a model that uses SVR for prediction~\citep{smola2004tutorial}.
The model was cross-validated to determine the best parameters using a grid
selection model. We evaluated linear and Radial Basis Function (RBF) kernels,
varying the regularization constant C between the values $1$, $10$, and $100$
and, for the RBF the $\gamma$ parameter was selected between $0.001$ and
$0.0001$.

\subsection{\emph{k}-Nearest Neighbors}

The \emph{k}-NN model is not traditionally a regression model. Therefore, we
made a simple adaptation to the algorithm to calculate the similarity between
two songs. Once a new query point was submitted, we found the \emph{k} nearest
neighbors and, between these neighbors, computed the mean value of them, and
that was the predicted value. The intuition between this heuristic is that
songs with similar features will tend to have similar scores.

\begin{table}
    \caption{Performance of the \emph{k}-NN model with the different models and with
    different number $k$ of neighbors. The best model is shown in
    \textbf{bold} face.}\label{tab:knn-performance}
    \begin{center}
    \begin{tabular}{lccc}
        \toprule
        & \multicolumn{3}{c}{$R^2$ score}\\
                            & 1 neighbor & 5 neighbors & 10 neighbors \\
        \cmidrule(r){2-4}
        Word2Vec raw        & 0.061      &  \textbf{0.273}    &  0.271 \\
        Word2Vec $s > 1\%$  & 0.239      &  0.177    &  0.158 \\
        Word2Vec $s > 2.5\%$& 0.146      &  0.084    &  0.085 \\
        \cmidrule(r){2-4}                              
        Tf-idf raw          & 0.322      &  0.296    &  \textbf{0.410} \\
        Tf-idf $s > 1\%$    & 0.166      &  0.251    &  0.235 \\
        Tf-idf $s > 2.5\%$  & 0.202      &  0.155    &  0.175 \\
        \bottomrule
    \end{tabular}
    \end{center}
\end{table}

\subsection{Linear Regression}

Linear regression is one of the simplest machine learning algorithms that most
often than not deliver good results. The algorithm minimizes the residual sum
of least squares between the observed responses in the dataset and the
responses predicted by linear regression. Due to this simplicity, this is
a model that must be evaluated. For if it can explain the data, Occam's razor
determines it should be selected as a good model.

\begin{table}
    \caption{Performance of the linear regression
    model.}\label{tab:lr-performance}
    \begin{center}
    \begin{tabular}{lcc}
        \toprule
                            & $R^2$ score & Accuracy \\
        \cmidrule(r){1-3}
        Word2Vec raw        & -0.010      & 0.000 (+/- 0.084) \\
        Word2Vec $s > 1\%$  & -0.018      & -0.023 (+/- 0.020)\\
        Word2Vec $s > 2.5\%$& -0.014      & -0.032 (+/- 0.060) \\
        \cmidrule(r){1-3}
        Tf-idf raw          &  0.012      & -0.193 (+/- 0.414)\\
        Tf-idf $s > 1\%$    &  0.024      &  0.015 (+/- 0.032)\\
        Tf-idf $s > 2.5\%$  & -0.005      & -0.038 (+/- 0.097) \\
        \bottomrule
    \end{tabular}
    \end{center}
\end{table}

\subsection{Approximate \emph{k}-NN with Locality Sensitive Hashing}

Building Locality Sensitive Hashing (LSH) forests~\citep{bawa2005lsh} is an
alternative when one is willing to trade accuracy for speed when doing nearest
neighbors search. Since we are already implementing \emph{k}-NN, it seems
appropriate to evaluate this algorithm as well, especially considering that
nearest neighbors search can become slow in problems of high dimensionality.
The performance of the LSH models is summarized in
Table~\ref{tab:lsh-performance}.

\section{Related Work}

\cite{eck2008automatic} use a set of boosted classifiers to map audio features
onto tags collected from the Web. Due to the nature of their classifier, it
uses the objective approach and, therefore, need the actual audio files, which
we are not using. \cite{berenzweig2004large} survey various music-similarity
measures and concludes that measures derived from co-occurrence in personal
music collections are the most useful ground truth metrics from those
evaluated. \cite{aucouturier2002music} define a song similarity measure based
on the analysis of songs' timbres, and also evaluate their metric.
\cite{johnson2014logistic} proposes a matrix factorization method that works
well for data with implicit feedback, such as song listening patterns.

\begin{table}
    \caption{Performance of the LSH model with the different models and with
    different number $k$ of neighbors. The best model is shown in
    \textbf{bold} face.}\label{tab:lsh-performance}
    \begin{center}
    \begin{tabular}{lccc}
        \toprule
        & \multicolumn{3}{c}{\shortstack{$R^2$ score\\for number of neighbors}}\\
                            & 1          & 5           & 10 \\
        \cmidrule(r){2-4}
        Word2Vec raw        &  0.296     &  0.189      &   \textbf{0.305} \\
        Word2Vec $s > 1\%$  &  0.234     &  0.133      &   0.167 \\
        Word2Vec $s > 2.5\%$&  0.105     &  0.096      &   0.069 \\
        \cmidrule(r){2-4}                                 
        Tf-idf raw          & -0.092     &  0.261      &   \textbf{0.301} \\
        Tf-idf $s > 1\%$    &  0.211     &  0.172      &   0.154 \\
        Tf-idf $s > 2.5\%$  &  0.042     &  0.097      &   0.074 \\
        \bottomrule
    \end{tabular}
    \end{center}
\end{table}

\section{Evaluation}

We have evaluated our system by sampling the information of 100.000 (a hundred
thousand) songs from our dataset. This was needed, since the full dataset
wouldn't fit the modest computers we had access to. This set was further
divided into two: a training test (which was also used for cross-validation)
and a test set, for evaluating the models' final performance.

In the initial phases of this work we had though about using the Root Mean
Squared Error (RMSE) function, defined below, for model performance, but recall
all similarity values are between zero and one.  Therefore, it would be hard to
get an intuitive feel of the model performance.

\begin{equation*}\label{eq:rmse}
    \mathrm{RMSE}(y, \hat{y}) = \sqrt{\frac{\sum_{i=1}^n{{(\hat{y}_i - y_i)}^2}}{n}}
\end{equation*}

Because of the previous discussion, we have decided to evaluate our models
using the coefficient of determination ($R^2$) score. The $R^2$ score is
defined below and its value is $1$ when the model can perfectly explain the
data and get only deviate below one. Notice that this allows the score to be
negative. Therefore, the more negative the $R^2$ score, the worse the model.
Another advantage of using $R^2$ is that, by definition, the $R^2$ score of
a predictor that always outputs the mean value of the dataset is zero. From
that it follows that models with $R^2$ values smaller than zero are not that
useful.

\begin{equation*}\label{eq:r2}
    \mathrm{R^2}(y, \hat{y}) = 1-\frac{\sum{{(y_i-\hat{y}_i)}^2}}{\sum{{(y_i-\bar{y})}^2}}
\end{equation*}

The results of the evaluation of the various models used in this work are shown
in Tables~\ref{tab:svm-feature-scaling}--\ref{tab:lsh-performance}. As can be
gathered, the best models were the ones based on the nearest neighbors models.
Also, notice that they perform significantly better than the predictor of the
mean. More striking is that the best results are found when the raw unfiltered
data is used, which is the exact opposite of the observed behavior of the SVR
model. The linear model stands between the SVR (the worse) and the \emph{k}-NN
models (the best), but it yields values too close to 0 to be particularly
useful, and a predictor that predicts the mean value of the data might be
better.

\section{Conclusion}

We have explored machine learning techniques for learning similarity between
songs. Particularly, we explored two different methods from the NLP field for
building feature matrices that were fed into the algorithms. Of these two
methods, the tf-idf one seems to give better results while also executing
faster than the Word2Vec one. We also selected models by means of cross-validation, 
splitting the data into a training and testing set, saving the
testing set only for the final evaluation.

About the learning algorithms themselves, it is interesting to notice that an
algorithm that generally performs very well in classification tasks (SVR) had
the worst performance with our dataset. More interesting is that a relatively
simple algorithm (\emph{k}-NN) that computes the mean of the query point's
neighbors had performance \emph{much} better than not only than the other
algorithms, but also of the estimator based on the mean (with $R^2$ score
zero).


\section{Lessons learned}

Most of the effort in preparing this paper was done in understanding and
adapting inconsistencies in the data obtained from the last\@.fm™ API\@. Also,
although data is abundant, and even though this is probably not considered big
data, the data is big enough to not fit in commodity computers. Still, the
lessons learned in this work allow for one approach for building the base of
recommendation systems.

\balance%

\bibliography{report}
\bibliographystyle{plainnat}

\end{document}